\documentclass[letterpaper, 10 pt, conference]{ieeeconf}  

\IEEEoverridecommandlockouts                              

\usepackage{cite}
\usepackage{amsmath}
\usepackage{amssymb}
\usepackage{graphicx}
\usepackage{subcaption}
\usepackage{multirow}
\usepackage[linesnumbered,ruled,vlined]{algorithm2e}
\usepackage{booktabs}
\usepackage{array}
\usepackage[hidelinks]{hyperref}

\usepackage{algpseudocode}
\makeatletter
\newcommand{\removelatexerror}{\let\@latex@error\@gobble}
\makeatother

\title{\LARGE \bf
RESC: A Reinforcement Learning Based Search-to-Control Framework for Quadrotor Local Planning in Dense Environments
}

\author{
     Zhaohong Liu, Wenxuan Gao, Yinshuai Sun, and Peng Dong
 \thanks{This work was supported by the National Natural Science Foundation of China under Grants 61803260. (\textit{Corresponding author: Peng Dong.})}%
 \thanks{The authors are with the School of Aeronautics and Astronautics, Shanghai Jiao Tong University, China
 (e-mail: zhliu25@sjtu.edu.cn; )}
\thanks{The source code will be released  at: \href{https://github.com/JaimeParker/resc-pilot}{https://github.com/JaimeParker/resc-pilot}}%
}

\begin{document}

\maketitle
\thispagestyle{empty}
\pagestyle{empty}

\begin{abstract}
Agile flight in complex environments poses significant challenges to current motion planning methods, as they often fail to fully leverage the quadrotor dynamic potential, leading to performance failures and reduced efficiency during aggressive maneuvers.
Existing approaches frequently decouple trajectory optimization from control generation and neglect the dynamics, further limiting their ability to generate aggressive and feasible motions.
To address these challenges, we introduce an enhanced Search-to-Control planning framework that integrates visibility path searching with reinforcement learning (RL) control generation, directly accounting for dynamics and bridging the gap between planning and control.
Our method first extracts control points from collision-free paths using a proposed heuristic search, which are then refined by an RL policy to generate low-level control commands for the quadrotor controller, utilizing reduced-dimensional obstacle observations for efficient inference with lightweight neural networks.
We validate the framework through simulations and real-world experiments, demonstrating improved time efficiency and dynamic maneuverability compared to existing methods, while confirming its robustness and applicability.
\end{abstract}

\section{Introduction}

Quadrotors are extensively used in a variety of applications, including rescue operations, fire and electricity inspection, and package delivery \cite{10478625}.
Achieving autonomous flight in complex environments necessitates effective motion planning, which is crucial for ensuring safe and efficient navigation.
Current state-of-the-art motion planning algorithms for quadrotors primarily focus on generating collision-free and dynamic feasible trajectories, subsequently optimizing them to achieve smoothness and aggressiveness \cite{ietgaofei}.
Time allocation plays a pivotal role in the trajectory generation process, as suboptimal time allocation can lead to inefficient trajectories and underutilization of the quadrotor dynamic capabilities \cite{timealloc}.

However, trajectory optimization methods often fail to fully leverage the quadrotor dynamic potential and struggle to operate effectively in high-agility scenarios \cite{rl-racing}.
Furthermore, the trajectories generated by these optimizers must be tracked by a controller, introducing delays, increased computational overhead, and mismatches between the optimized and actual flight trajectories, particularly under aggressive maneuvers.

\begin{figure}[t]
    \centering
    \includegraphics[width=0.95\linewidth]{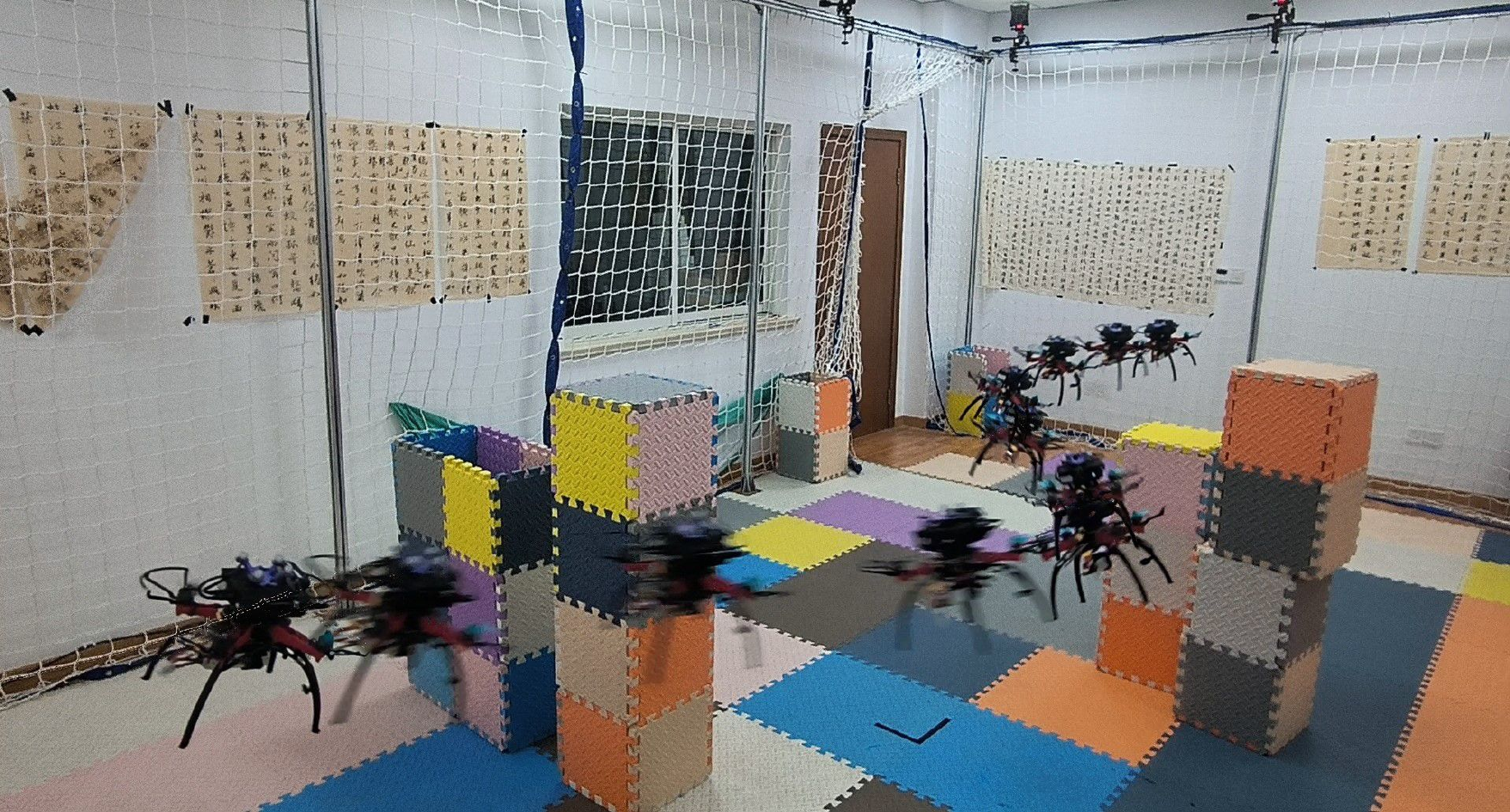}
    \caption{Our quadrotor flying in challenging indoor environments while avoiding obstacles.}
    \label{fig:realworld-1}
\end{figure}

\begin{figure}[t]
    \centering
    \includegraphics[width=0.9\linewidth]{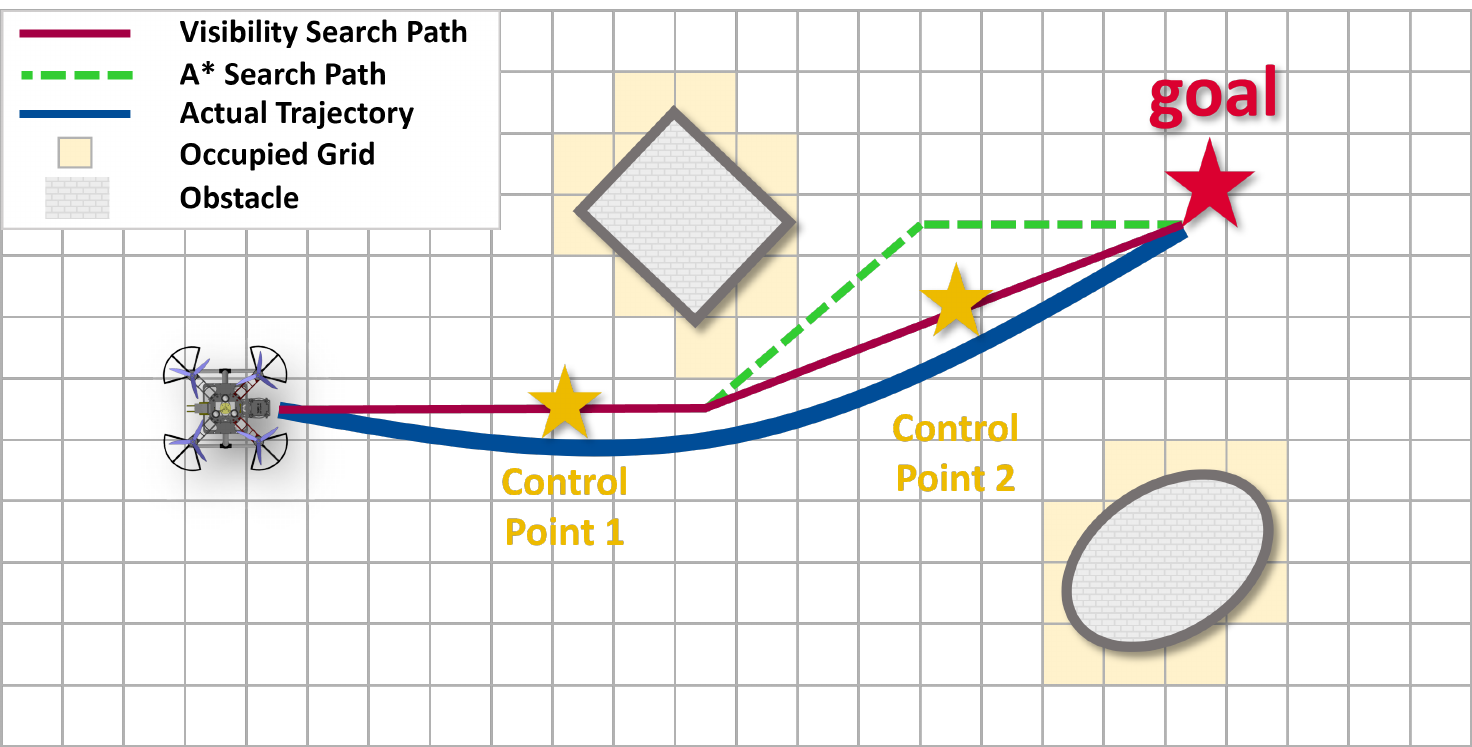}
    \caption{Illustration of the proposed method. RL policy is used to generate control commands based on the quadrotor state, obstacles, and control points.}
    \label{fig:intro}
\end{figure}

To address these challenges, we propose an improved Search-to-Control RL-based motion planner for quadrotor autonomous flight in dense environments.
Our approach aims to maximize the utilization of the quadrotor dynamic capabilities while relying solely on a rate controller.
Our method begins with a path searching algorithm based on heuristic search, which constructs a collision-free, minimum-length path in a discrete grid space (Fig. \ref{fig:intro}).
This high-level path is then converted into control points, which serve as input for the RL control generator.
The RL policy network processes observations of the quadrotor state, control points, and environmental information to generate low-level control commands.
Finally, the generated commands are sent directly to the quadrotor rate loop controller, enabling the execution of agile and precise motions.

In this letter, we design a \textbf{R}einforcement L\textbf{E}arning based \textbf{S}earch-to-\textbf{C}ontrol framework called \textbf{RESC}.
Compared to existing motion planners, our method departs from trajectory generation and optimization frameworks by directly generating control commands using a RL policy.
This approach can handle discrete paths that may not inherently satisfy kinodynamic and non-holonomic constraints, producing aggressive, collision-free motions within specified dynamic limits, without relying on iterative numerical optimization.
We evaluate the effectiveness and robustness of the proposed method through extensive simulations and real-world experiments across diverse scenarios.
Our key contributions are summarized as follows:

\begin{itemize}
    \item[(1)] A unified Search-to-Control planning framework that combines visibility-based path searching with RL-driven control generation, enabling aggressive and feasible motions without explicit trajectory representation or optimization. 
    \item[(2)] A comprehensive RL environment that fully utilizes quadrotor dynamic potential to train policies, facilitating the generation of low-level control commands for diverse environments and seamless integration with the pathfinding module.
    \item[(3)] A reduced-dimensional obstacle observation method that efficiently captures environmental complexity, enabling lightweight neural network training and inference for robust control generation.
    \item[(4)] Extensive simulation and real-world experiments on an autonomous quadrotor, demonstrating the robustness and practical applicability of the proposed framework. The source code is released for further research.
\end{itemize}

\section{Related Work}

\subsection{Trajectory Optimization Methods}
Trajectory optimization methods prevail as the predominant approach for quadrotor motion planning,
which focus on generating smooth, dynamically feasible, and collision-free trajectories by modeling as optimal control problems.
The pioneering approach is the minimum-snap trajectory generation\cite{minimumsnap} in differential flatness outputs, representing the trajectory as a polynomial and using gradient descent to iteratively modify the time segments.
Several works propose two-stage solving methods where trajectory generation occurs in the second stage \cite{safecorridor,gradbased,bsplineframework,fastmarching,fastplanner,faster}.
For instance, a safe flight corridor composed of convex polyhedra is introduced in \cite{safecorridor}, providing constraints for optimization. 
Additionally, B-splines are utilized for kinodynamic path searching in \cite{bsplineframework}, where an elastic optimization method is employed to refine trajectories.
The Euclidean Signed Distance Field (ESDF) map is employed in \cite{eth2016} to obtain collision potentials and in \cite{fastmarching} to find a path in the velocity field, improving time allocation.
\cite{fastplanner} proposes a comprehensive pipeline for quadrotor motion planning, including kinodynamic path searching, B-spline trajectory generation, time allocation, and nonuniform B-spline optimization based on ESDF.
Subsequent works \cite{fastplanner2} and \cite{raptor} introduce topological path and risk-aware yaw planning.
Given the computational intensity of constructing ESDF, \cite{egoplanner} proposes an ESDF-free gradient-based planner.
Wang \textit{et al.} \cite{minco} focuses on spatial-temporal deformation, which aligns with our approach by considering both geometric and temporal planning concurrently.
These methods effectively generate smooth, safe, and feasible trajectories. 
However, they may be conservative regarding the ignorance of quadrotor dynamic capabilities.

\subsection{RL based Methods}
RL-based methods have gained significant traction in quadrotor motion planning due to their capability to optimize non-convex and discontinuous objectives.
In \cite{sci-wildflight}, the authors propose an imitation learning (IL)-based teacher-student pipeline that directly maps noisy sensory depth images to polynomial trajectories.
While IL is quite efficient, its scalability is limited by the need for expert data collection.
The work in \cite{rl-mintimeflight} utilizes a topological path-searching algorithm to generate guiding paths, which are subsequently refined by reinforcement learning to produce control commands.
This two-stage method performs well in selected scenarios, outperforming some traditional methods.
Song \textit{et al.} \cite{rl-pawareflight} extracts features from depth images and uses a teacher-student IL framework similar to \cite{sci-wildflight}. 
Furthermore, it generates control commands directly and includes a perception reward to guide yaw planning.
In \cite{sci-rlracing, rl-racing}, Song \textit{et al.} investigates the strengths and limitations of optimal control and RL, proposing that RL is advantageous as it can optimize better objectives that may be infeasible for optimal control methods to address.
In \cite{10582409}, Zhao \textit{et al.} propose a hierarchical RL framework where an outer-loop policy dynamically configures speed and acceleration constraints based on online observations, significantly improving flight efficiency and safety in cluttered environments.

However, these methods frequently rely on imitation learning (IL) with expert knowledge or are trained in fixed environments, limiting their generalization and transferability to unknown environments, thereby hindering autonomous flight in new scenarios.
Instead of using RL for the entire planning process, we employ it as an optimizer to refine control points and directly generate control commands.
This strategy allows for more efficient learning and better adaptation to complex environments.

\section{System Overview}

\subsection{Planning Structure}

\begin{figure}[t]
    \centering
    \includegraphics[width=0.98\linewidth]{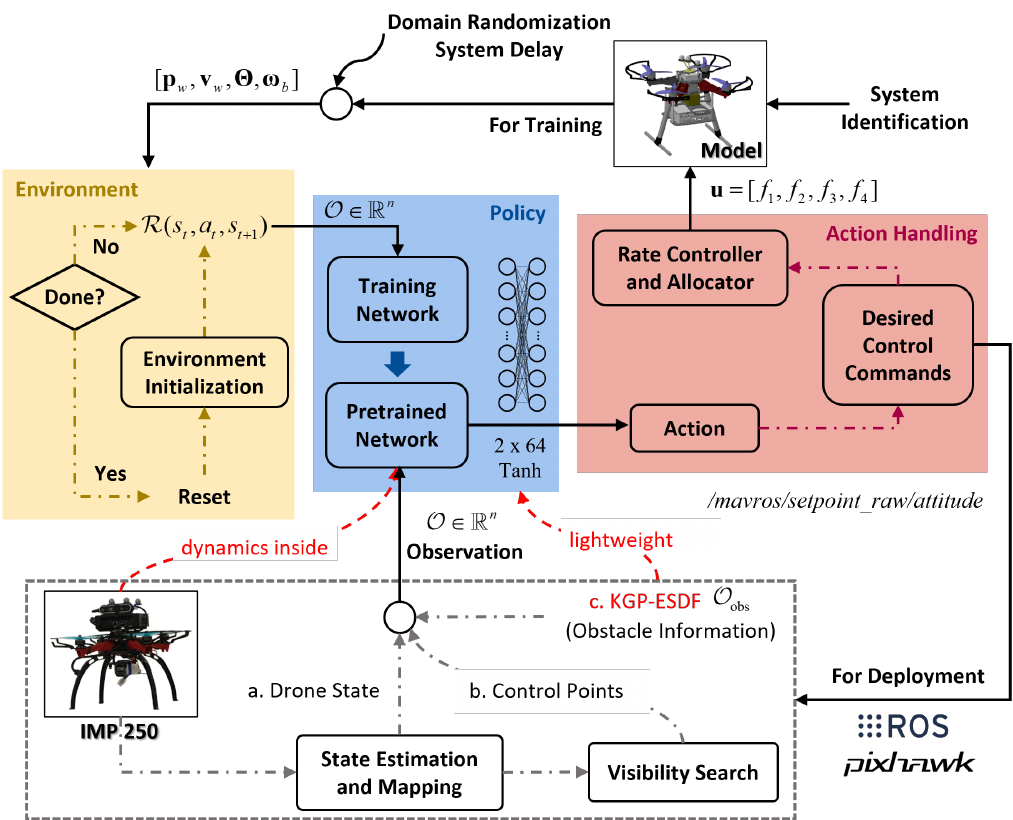}
    \caption{An overview of our proposed framework. It focuses exclusively on control generation during training. During deployment, it integrates state estimation and mapping data to generate control commands through a two-step process.}
    \label{fig:framework}
\end{figure}

The proposed framework, illustrated in Fig. \ref{fig:framework}, follows a planning process that consists of two stages.
First, the Visibility Path Searching algorithm constructs a reference path from the start to the goal, ensuring obstacle avoidance and path length minimization. 
This reference path is discretized into control points based on the current position and constraints among control points (Section \ref{sec:pp}).
Second, the RL Control Generation method is employed to generate control commands that optimize the quadrotor motion.
These commands guide the quadrotor to follow the control points within a specified tolerance, determining the proximity requirements and strategies for obstacle avoidance in continuous space, as detailed in Section \ref{sec:rl}.

\subsection{Quadrotor dynamics}
The quadrotor is modelled as a rigid body governed by a linear system with decoupled translational and rotational dynamics.
The control input vector $\boldsymbol{u}\in \mathbb{R}^4$ corresponds to the individual rotor thrusts, specifically defined as $\boldsymbol{u}=[f_1, f_2, f_3, f_4]^T$ with $f_i$ denotes the thrust scalar of the $i$-th rotor.
The complete state of the quadrotor is defined as $\boldsymbol{x}=[\boldsymbol{p}_w, \dot{\boldsymbol{p}_w}, \ddot{\boldsymbol{p}_w}, \boldsymbol{\Theta}, \boldsymbol{\omega}_b]$ , where $\boldsymbol{p}_w \in \mathbb{R}^3$ is the position in world frame, $\boldsymbol{\Theta}=\left[ \phi, \theta, \psi \right]^T \in \mathbb{SO}^3$ represents roll, pitch and yaw angle, and $\boldsymbol{\omega}^b$ denotes angular velocity in body frame.

The dynamics equations are 

\begin{equation}
    \ddot{\boldsymbol{p}}_w = -g \boldsymbol{e}_3 + \sum_{i=0}^{3}\frac{f_i}{m} \boldsymbol{R}_b^w \boldsymbol{b}_3.
\end{equation}

\begin{equation}
    \boldsymbol{J}\boldsymbol{\dot{\omega}}^b=-\boldsymbol{\omega}^b \times\boldsymbol{J}\boldsymbol{\omega}^b + \boldsymbol{G_a}+\boldsymbol{\tau}.
\end{equation}

\noindent where $\boldsymbol{v}_w$ is velocity, $\boldsymbol{R}_b^w$ is rotation matrix from body frame to world frame, $\boldsymbol{b}_3$ is z axis in body frame, $f_i$ is thrust of each rotor, $m$ is mass, $\boldsymbol{J}$ is diagonal inertia matrix, , $\boldsymbol{G_a}$ and $\boldsymbol{\tau}=[\tau_x, \tau_y, \tau_z]^T$ is gyroscopic torque and torque by thrusts respectively.
We use Euler angles to represent rotation and set a limit on roll and pitch angles to avoid singularity.

The control allocation matrix that can be used to get relation between thrusts and torques is given by

\begin{equation}
    \begin{bmatrix}
        1& 1& 1& 1\\
        -l/\sqrt{2}& -l/\sqrt{2}& l/\sqrt{2}& l/\sqrt{2}\\
        -l/\sqrt{2}& l/\sqrt{2}& l/\sqrt{2}& -l/\sqrt{2}\\
        -\kappa& \kappa& -\kappa& \kappa
    \end{bmatrix}
    \boldsymbol{u}
    =
    \begin{bmatrix}
        f_T \\ \tau_x \\ \tau_y \\ \tau_z
    \end{bmatrix}.
\end{equation}

\noindent where $\kappa$ denotes rotor's torque coefficient, and $l$ represents the distance from the center of mass to the rotor center.

\section{Visibility Path Searching}\label{sec:pp}
Our path searching algorithm combines elements of the A*, jump point search (JPS) \cite{JPS}, and visibility graph methods.
We inherit the heuristic function and the open and closed list architecture from the A* algorithm but implement an alternative cost function. 
Elements of the JPS method are incorporated to reduce the number of nodes in the open list, minimizing computational costs associated with sorting these nodes. 
Additionally, we leverage the optimal distance characteristics of the visibility graph method without constructing complex polygon edges and vertices.

The proposed strategy facilitates omnidirectional movement through obstacle corner detection and visibility checks, rather than restricting movement to increments of $\pi/4$ in 2D plane.
This approach is implemented by selecting obstacle corners in the \textbf{Corner}$\left(\right)$ function and assessing voxel visibility in the \textbf{VisibleCheck}$\left(\right)$ function. 
As a result, this method reduces the computational and memory overhead typically required for accessing each vertex and edge of the voxels.
Subsequently, the corners are linked to form paths, which are then sampled to generate control points.
The detailed algorithm is presented in Algorithm \ref{algo:ps}.

\begin{algorithm}[t]
    \SetKwInOut{Input}{input}\SetKwInOut{Output}{output}
    
    \Input{2D ESDF map, start $\mathcal{P}_s$ and goal $\mathcal{P}_g$}

    \While{not $\mathcal{O}$.\bf{empty}$\left(\right)$}{
        $n_c \leftarrow \mathcal{O}$.\textbf{pop}$\left(\right)$, $\mathcal{C}$.\textbf{emplace}$(n_c)$ \;
        \If{\bf{RandomVisibleCheck}$(n_c, \mathcal{P}_g)$}{
            \Return{{\bf{TraversePath}}$\left(\right)$} \;
        }
        \For{$r \in r_{\text{max}}$}{
            $n_h \leftarrow${\bf{GetNeighbors}}$\left(r, n_c\right)$ \;
            \If{$\left(n_h=\mathcal{P}_g\right) \cap $\bf{VisibleCheck}$\left(n_c,n_h\right)$}{
                \Return{{\bf{TraversePath}}$\left(\right)$} \;
            }
            \If{\bf{Corner}$\left(n_h\right) \cap$\bf{VisibleCheck}$\left(n_c,n_h\right)$}{
                $g_{temp} \leftarrow n_c.g_c +${\bf{EuclideanCost}}$\left(n_c,n_h\right)$ \;
                \If{$\mathcal{O}.${\bf{contain}}$\left(n_h\right) \cap \left(n_h.g_h < g_{temp}\right)$}{
                    {\bf{continue}}\;
                }
                $n_h.{parent}\leftarrow n_c$, $n_h.g_h \leftarrow g_{temp}$ \;
                $n_h.f \leftarrow g_{temp} + ${\bf{Heuristic}}$\left(n_c,n_h\right)$ \;
                $\mathcal{P}.${\bf{emplace}}$\left(n_h\right)$ \;
            }
        }
    }

    \caption{Visibility Path Searching.}\label{algo:ps}
\end{algorithm}

\subsection{Obstacle Corner Search}
We define an obstacle corner as a voxel with only one occupied neighbor, indicating its location at the edge of an obstacle.
In the context of the Visibility Graph, the optimal path from the start to the goal point traverses vertices that represent the corners of obstacles.
Therefore, obstacle corners \textbf{Corner}$\left(\right)$, including the start and goal points, are exclusively utilized as elements in the open list.
This approach reduces the number of nodes in the open list and decreases sorting costs, similar to JPS, while maintaining the optimal path characteristics of the Visibility Graph.

The cost of a obstacle corner using \textbf{EuclideanCost}$\left(\right)$ is defined as

\begin{equation}
    g(n) = g(n-1) + ||\boldsymbol{p}_n - \boldsymbol{p}_{n-1}||.
\end{equation}

\subsection{Visibility Expansion}
When a new obstacle corner is detected, the Bresenham algorithm\cite{bresenham} is employed to perform a visibility check using \textbf{VisibleCheck}$\left(\right)$, determining whether the corner is visible from the current node. 
Only corners confirmed to be visible are considered as child nodes and candidates for the open list, ensuring that a connected path of vertices via line segments is traced from start to goal.
We adopt an expansion method from Hybrid A*\cite{hybridAstar} in \textbf{RandomVisibleCheck}$\left(\right)$ to verify the existence of a collision-free path between the current node and the goal along a line segment. 
The invocation frequency of this function increases as the distance to the goal decreases.


\section{RL Based Control Generation}\label{sec:rl}

\subsection{Policy Framework}
We frame the control point optimization and control generation problem within a reinforcement learning (RL) framework.
The quadrotor agent selects actions based on the policy $\pi(\boldsymbol{a}_t|\boldsymbol{s}_t)$, transitioning from state $s_t$ to $s_{t+1}$ according to the dynamics-kinematics update of the system.
The agent receives a reward $\mathcal{R}(s_t, a_t, s_{t+1})$ that guides learning.
Detailed configurations of the state space, action space, and reward function will be discussed in subsequent sections. 

\subsubsection{Observation and action space}
The observation space contains three main components: the state of the quadrotor, control points and obstacle information.
The state vector is defined as

\begin{equation}
    \boldsymbol{s} = \left[ \boldsymbol{p}_w(t), \boldsymbol{v}_w(t), \boldsymbol{\omega}_b(t), \boldsymbol{\Theta}(t), \boldsymbol{p}_{\text{cp}},  \boldsymbol{O}_{\text{obs}} \right],
\end{equation}

\noindent where $\boldsymbol{p}_{\text{cp}}$ represents control points, and $\boldsymbol{O}_{\text{obs}}$ denotes obstacle information, .

The observation vector is derived from the state vector, directly incorporating components such as $\boldsymbol{v}_w(t)$ , $\boldsymbol{\omega}_b(t)$, $\boldsymbol{\Theta}(t)$, and $\boldsymbol{O}_{\text{obs}}$.
Other elements of the state are converted into lower-level features, making them more generalized and easier for policy training.
For example, we use relative position from quadrotor position to control points instead of absolute position.
Additionally, a normalized vector representing the x-axis direction of body frame in world frame ($x_b^w$) is used to guide the yaw to stay visiualized for control point based on the field-of-view (FOV) of depth camera.
Obstacle observation $\boldsymbol{O}_{\text{obs}}$ will be discussed in the following section.

The action produced by RL policy is $\boldsymbol{a}(t)=\left[ f_T, \boldsymbol{\omega}_b^{req} \right]$, where $f_T$ is the collective thrust and $\boldsymbol{\omega}_b^{req}$ represents the desired body rate.
This control approach has been proved effective for learning-based quadrotor control and planning tasks \cite{benchmark-action}.
Although commonly used in drone racing tasks due to its low latency and high frequency, it can present a gap between simulation and real-world performance.
Despite this limitation, this control approach leads to smoother acceleration and velocity, improving energy efficiency and safety compared to higher-dimensional control approaches.

\subsubsection{Reward function}
According to paper\cite{sci-rlracing}, RL outperforms Optimal Control (OC) in terms of optimizing a better objective that OC may not be able to solve.
From this perspective, we design a reward function that explicitly represents agile motion, collision avoidance, and smoothness in control.
The reward function is defined as

\begin{equation}
    r(t) = k_p r_p + r_c + k_d r_d + k_v r_v + k_s ||\boldsymbol{\omega}_b|| + r_f,
\end{equation}

\noindent where $r_p$ denotes the customized progress reward, represents the penalty for collisions, $r_d$ is the penalty for dynamic violations, $r_v$ is the penalty for FOV violations, and $r_f$ rewards task completion.

Specifically, the progress reward $r_p(t)$ is defined as

\begin{equation}
    r_p(t) = \frac{||\boldsymbol{p}_{\text{int}}-\boldsymbol{p}_{w, i-1}||-||\boldsymbol{p_{\text{int}}}-\boldsymbol{p}_{w,i}||}{v_{\text{max}}\Delta t} + r_t,
\end{equation}

\noindent where $\boldsymbol{p}_{\text{int}}$ denotes the intersection point if the vector $\boldsymbol{p}_{w, i} - \boldsymbol{p}_{w, i-1}$ intersects the control point area; otherwise, the intersection point is directly $\boldsymbol{p}_{\text{cp}, i}$.
We normalize the progress reward and subsequently introduce a negative term $r_t$, thereby imposing a time penalty which maximizing the utilization of the quadrotor dynamic capabilities.

For dynamic violation penalty $r_d$, we impose limits on acceleration and velocity to ensure dynamic smoothness and safety.
Additionally, attitude limits are set to prevent overly aggressive maneuvers and avoid singularities.

\begin{equation}
    r_d(t) = -n, \quad n \in \{0, 1, 2, 3\},
\end{equation}

\noindent where $n$ represents the number of elements from the set $\left[ \boldsymbol{v}_w(t), \boldsymbol{a}_w(t), \boldsymbol{\Theta}(t) \right]$ that exceed their prescribed limits.

The FOV violation penalty $r_v$ is set to $-1$ if neither of the next two control points is within the FOV of the depth camera.
This ensures that the quadrotor maintains visibility of the control points.

\subsection{Training methodology}

\begin{figure}[t]
    \centering
    \includegraphics[width=0.95\linewidth]{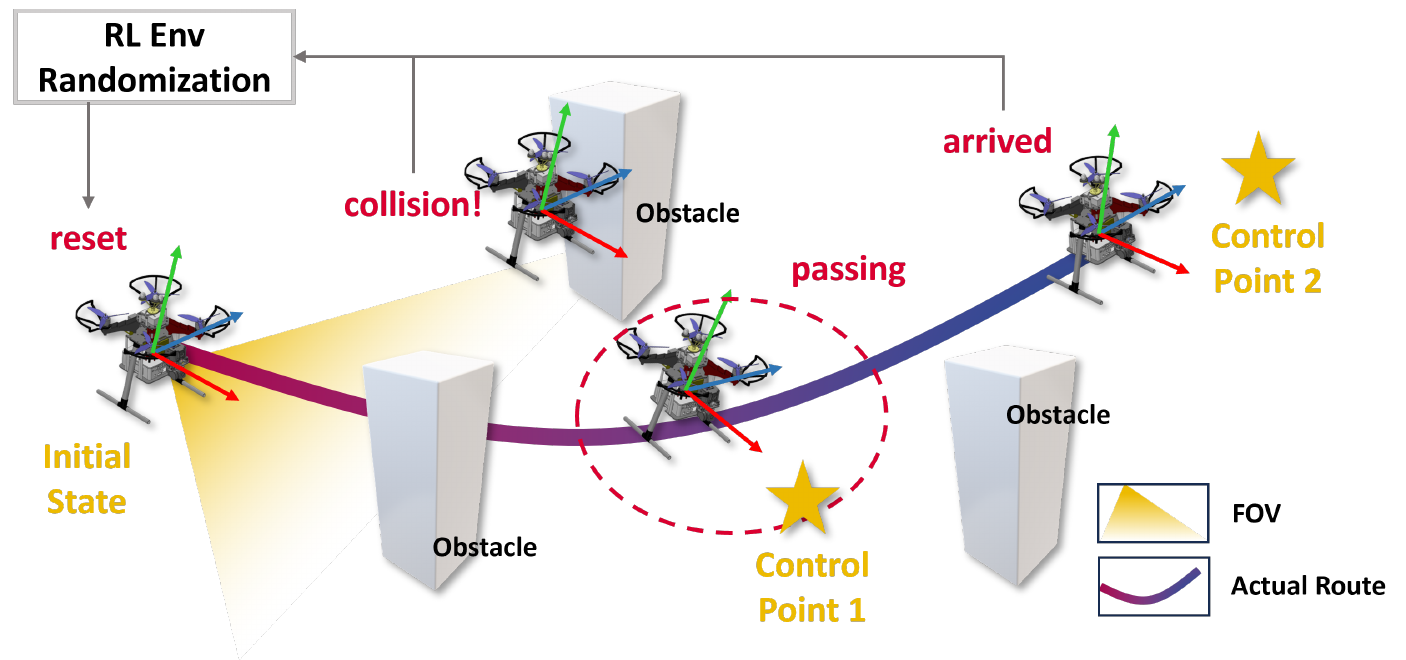}
    \caption{Visualization of our RL training environment. The environment is reset when collision or the last control point is reached.}
    \label{fig:rlscene}
\end{figure}

We create a training environment based on OpenAI gym\cite{openaigym}, providing an interface for creating a customized environment.
At the beginning of each episode, the velocity, body rate, orientation, control points and obstacles are randomly initialized following a predefined probabilistic pattern, while the position if fixed at the center of the map (Fig. \ref{fig:rlscene}).

\subsubsection{Environment randomization}
The environment is initialized with control points and obstacles, with the quadrotor starting at the map center. 
The first control point is randomized within a specified distance from the quadrotor position. 
Subsequent control points are placed within a defined distance and angle from the previous one.
Velocity and orientation constraints are applied to keep the quadrotor motion stable, ensuring its velocity remains basically towards the first control point, and its pitch, roll, and yaw stay within specified limits.

Obstacles are placed along or near the lines connecting the quadrotor to the control points, ensuring collision-free paths while allowing some proximity to obstacles, rather than creating a maze-like environment.

\subsubsection{Control point observation}

Following Song et al. \cite{rl-racing}, we observe two relative positions of the control points.
For the last control point, the relative position of the quadrotor to this point is repeated.
We define a finite-height column with a radius of $d_{\text{hor}}$ and a height of $2d_{\text{hgt}}$ to represent the area within which the control point is considered passed.
The observation is then updated to the next two control points.
With this design and the randomization of the environment, the use of a small number of control points (we use two in our training) enables efficient training while allowing generalization to more complex tasks with a larger number of control points.

\subsubsection{Obstacle observation}
\begin{figure}[t]
    \centering
    \includegraphics[width=0.9\linewidth]{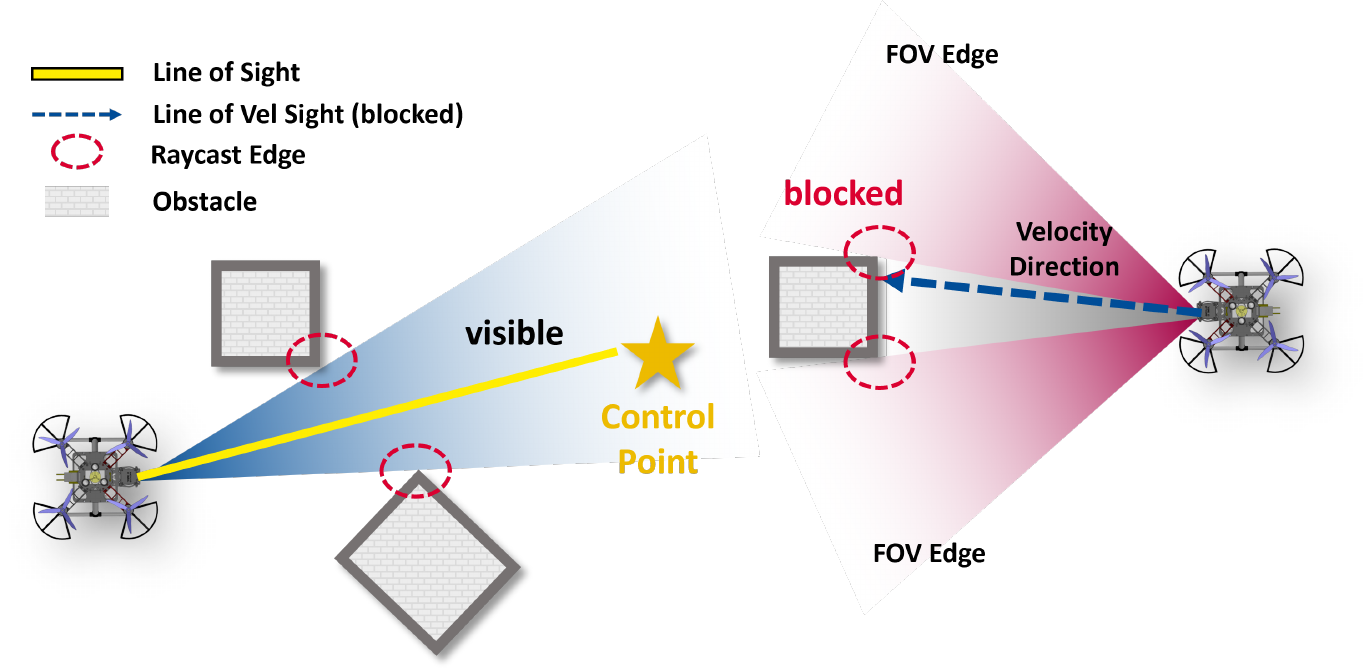}
    \caption{Example of Kinematic Pseudo-Raycast. 
    The blue region represents the guided raycast from the quadrotor position to the control point, while the red region indicates the guided raycast from the quadrotor position to the velocity control point.}
    \label{fig:kgp}
\end{figure}

We introduce a novel method to describe obstacle information efficiently, called Kinematic Guided Pseudo-Raycast and ESDF Perception (KGP-ESDF), as shown in Fig. \ref{fig:kgp}.

The core concept of KGP is to identify the most relevant obstacle edges for the quadrotor motion towards the control point, simultaneously avoiding collisions along the current velocity direction.
First, the algorithm performs a sector-based search along the line from the current position $\boldsymbol{p}_{w}$ to the control point to identify the nearest obstacle edges, denoted as $\boldsymbol{o}_{\text{cp}}$.
Subsequently, a similar search is conducted along the line from $\boldsymbol{p}_{w}$ to the velocity control point $\boldsymbol{p}_{\text{vcp}}$ as defined by the equation

\begin{equation}
    \boldsymbol{p}_{vcp} = \boldsymbol{p}_w + \frac{\boldsymbol{v}_w}{||\boldsymbol{v}_w||} \cdot ||\boldsymbol{p}_w - \boldsymbol{p}_{\text{cp}, i}||,
\end{equation}

\noindent to identify the nearest obstacle edges, denoted as $\boldsymbol{o}_{\text{vcp}}$.

The relative positions of these edge points are incorporated as part of the observation.
Additionally, we record the distance $d_{\text{vel}}$ from the current position to the velocity control point, which represents the collision-free distance.

We generate $\boldsymbol{o}_{\text{sdf}}$ to represent the surrounding obstacle information, derived from the ESDF map.
Finally, the obstacle observation $\boldsymbol{O}_{\text{obs}}$  is constructed as

\begin{equation}
    \boldsymbol{O}_{\text{obs}} = \left[ \boldsymbol{o}_{\text{cp}}, \boldsymbol{o}_{\text{vcp}}, d_{\text{vel}}, \boldsymbol{o}_{\text{sdf}} \right].
\end{equation}

\subsection{Sim-to-real transfer}

Transferring a policy trained in a customized environment to simulation and then to a real-world platform is challenging due to factors like inaccurate modeling of quadrotor dynamics, system delays, low-level controller errors, and action oscillations introducing high-order system complexities. 
We implement strategies to mitigate these issues.

We first train the policy in a customized Gym-based environment before transferring it to Gazebo simulation with PX4 firmware and then testing it on our quadrotor platform without fine-tuning.
We meticulously identified the Iris drone and our quadrotor models, achieving a relatively accurate representation.
To ensure compatibility with low-level control, we utilize the PX4 multicopter rate controller and control allocator to compute thrusts from the desired body rate and collective thrust outputs of the RL policy.

Additionally, we apply domain randomization and action delay queue during RL training to enhance policy robustness and prevent overfitting, compensating for numerical integration errors and modeling discrepancies between simulation and reality.
We randomized the quadrotor physical properties, including mass, inertia, and thrust mapping, and ignored battery voltage drop and motor heating effects due to short flight duration.
 
\section{Experiment Results}

\begin{figure*}[t]
    \centering
    \includegraphics[width=0.9\linewidth]{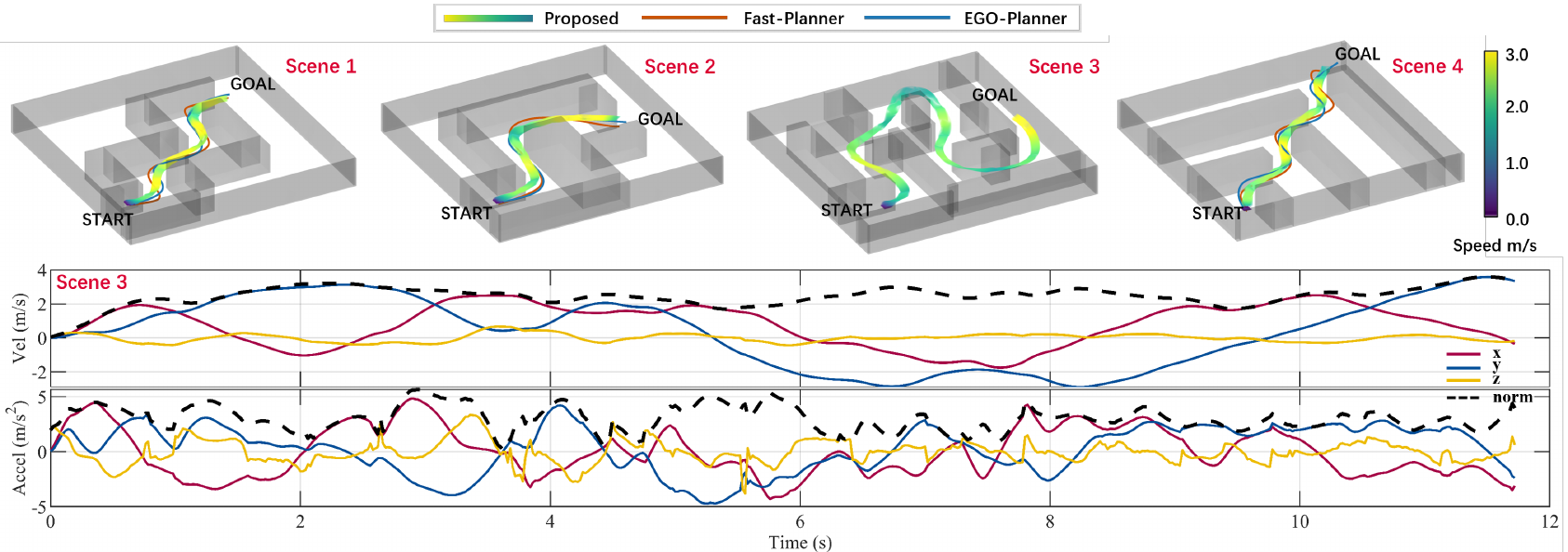}
    \caption{Visualization of the planning comparison in 4 scenes between baseline and proposed methods.}
    \label{fig:comparison}
\end{figure*}

\begin{table*}[t]
    \caption{Planners comparison in four scenes with respect to time span and energy consumption.}
    \begin{center}
        \begin{tabular}{c|>{\centering\arraybackslash}p{1cm}>{\centering\arraybackslash}p{1cm}>{\centering\arraybackslash}p{1cm}>{\centering\arraybackslash}p{1cm}|>{\centering\arraybackslash}p{1cm}>{\centering\arraybackslash}p{1cm}>{\centering\arraybackslash}p{1cm}>{\centering\arraybackslash}p{1cm}}
            \toprule
            & \multicolumn{4}{c|}{Time Span ($s$)} & \multicolumn{4}{c}{Energy ($m^2/s^5$)} \\
            \midrule
            Method & scene1 & scene2 & scene3 & scene4 & scene1 & scene2 & scene3 & scene4 \\
            \midrule
            Fast-Planner\cite{fastplanner} & 7.43 & 6.03 & $\inf$ & 5.33 & 176.55 & 44.36 & 0 & 151.81 \\
            EGO-Planner\cite{egoplanner} & 4.64 & 4.88 & $\inf$ & 4.24 & 824.03 & 497.84 & 0 & 578.94 \\
            RESC & \textbf{4.5} & \textbf{4.48} & \textbf{11.7} & \textbf{4.00} & 697.11 & 682.53 & 1147.1 & 611.52 \\
            \bottomrule
        \end{tabular}
    \end{center}
    \label{tab:comparison}
\end{table*}

\subsection{Training details}

The policy is trained using the Proximal Policy Optimization (PPO)\cite{PPO} algorithm in Stable-Baselines3\cite{stable-baselines3}, which has demonstrated strong performance in various continuous control tasks, including drone racing. 
We train two separate models for scenarios with and without obstacles to streamline training and improve efficiency.
This approach also mitigates the issue of forgetting, which we encountered when sequentially training a single model first on obstacle-free and then on obstacle-present cases.
The decision to use a specific model depends on the values from the ESDF map and the camera's ideal range.

The quadrotor physical properties and the RL algorithm's hyperparameters are listed in Table \ref{tab:quadandrl}. 
Following previous research, we employ Runge-Kutta 4th order integration to update the quadrotor state, excluding aerodynamic drag during training due to its insignificance for the current $v_{\text{max}}$.

All policies are trained on a desktop computer equipped with an NVIDIA RTX 4090 GPU and an Intel i9-13900K CPU.

\begin{table}[t]
    \caption{Parameter of our method}
    \begin{center}
        \begin{tabular}{ccccc}
            \toprule
            & Variable & Value & Variable & Value\\
            \midrule
            \multirow{4}{*}{IMP250} 
            & $m$ [kg] & 1.64 & $\kappa$ [-]& 0.012 \\
            & $i_{\text{xx}}$ [$\text{g m}^{-}$] & 0.011 & $l_{\text{x}}$ [m] & 0.88 \\
            & $i_{\text{yy}}$ [$\text{g m}^{-}$] & 0.010 & $l_{\text{ys}}$ [m] & 0.88 \\
            & $i_{\text{zz}}$ [$\text{g m}^{-}$] & 0.007 & $l_{\text{yl}}$ [m] & 0.88 \\

            \hline
            \multirow{9}{*}{RL} 
            & $a_{\text{max}}$ [$\text{m s}^{-2}$] & 6.0 & $v_{\text{max}}$ [$\text{m s}^{-1}$] & 3.0 \\
            & $k_{p}$ [-] & 2.0 & $k_{d}$ [-] & 5.0 \\
            & $k_v$ [-] & 5.0 & $k_s$ [-] & 0.01 \\
            & $r_c$ [-] & -600 & $r_f$ [-] & 300 \\
            & $d_{\text{hor}}$ [m] & 0.5 & $d_{\text{hgt}}$ [m] & 0.2 \\
            & $\Delta t$ [s] & 0.02 & $r_t$ [-] & -1.5 \\
            & $n_{\text{steps}}$ [-] & 2048 & $\gamma$ [-] & 0.99 \\
            & batch size [-] & 128 & learning rate [-] & 0.0003\\
            & $n_{\text{epochs}}$ & 10 & $\lambda$ (GAE) & 0.95\\

            \bottomrule
        \end{tabular}
    \end{center}
    \label{tab:quadandrl}
\end{table}

\subsection{Analysis and Comparison}

We compare our method with Fast-Planner \cite{fastplanner} and EGO-Planner \cite{egoplanner} in terms of time to reach the same position in four different scenes(Fig. \ref{fig:comparison}).
To ensure fairness, we remove control disturbances and initialize both methods from the same state.
The baseline methods use the same maximum velocity ($3m/s$), acceleration ($6m/s^2$), and inflated obstacle size ($0.3m$) as our method.
Additionally, we impose stricter limits on z-axis position ($1 \pm 0.3m$), sensing horizon ($5.0m$), and searching horizon ($7.5m$) to align with our method.
Other parameters are set to default values.
The comparison of baseline methods and our method is shown in Table \ref{tab:comparison}.
In scenes 1, 2, and 4, our method outperforms the baseline approaches in terms of time efficiency, demonstrating its ability to generate more aggressive control commands. 
The drone achieves the desired motion accurately after executing the control commands, benefiting from the integration of dynamics in our policy network.
In scene 3, only our method successfully reaches the goal, while the baseline methods fail to achieve the same result under the given acceleration and velocity constraints.

Consequently, our method shows better performance in terms of time efficiency, although it results in higher energy consumption compared to the baseline methods, primarily due to the aggressive control commands generated by the RL policy.

\begin{figure*}[t]
    \centering
    \includegraphics[width=0.9\linewidth]{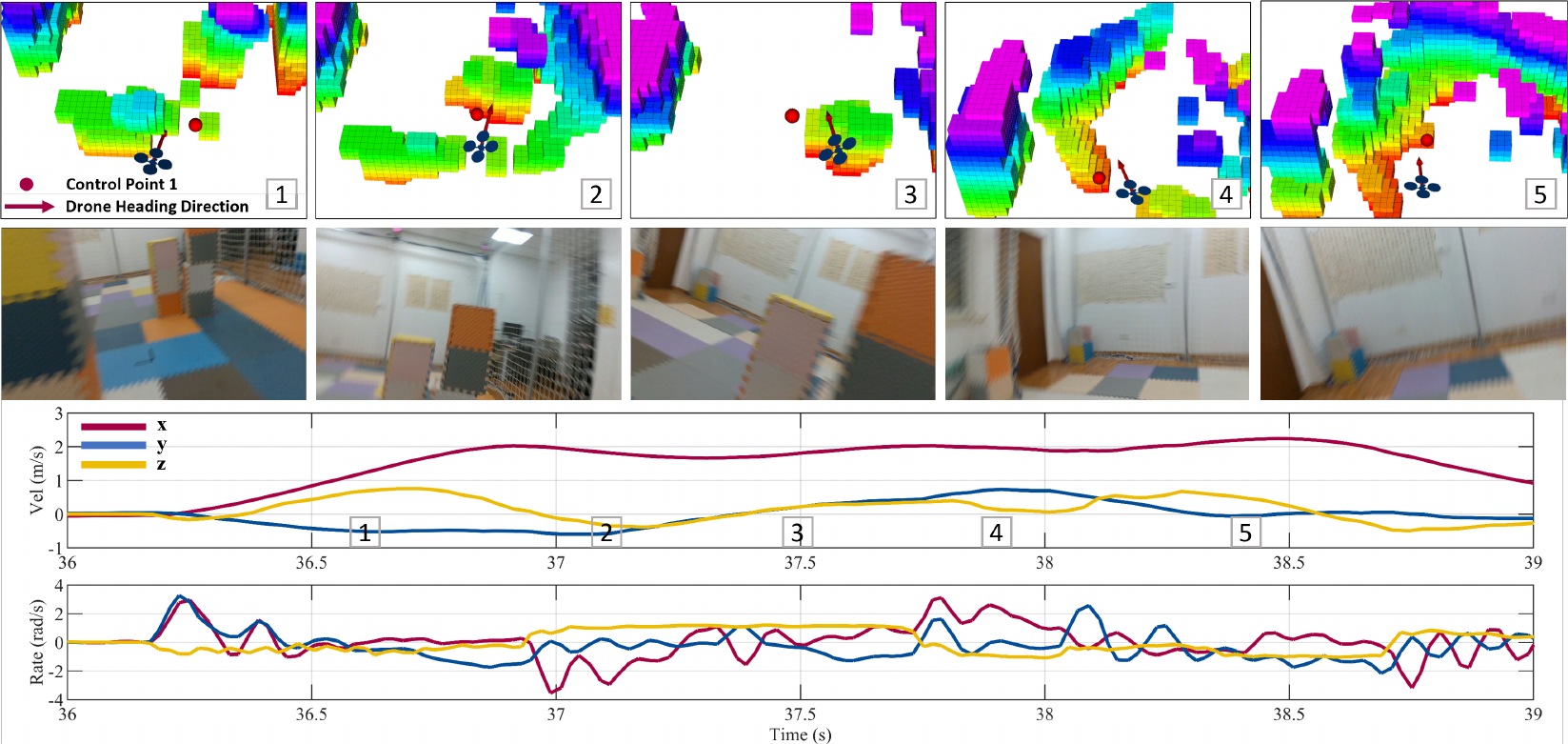}
    \caption{Visualization of a indoor planning experiment over a short period of time with actual velocity and body rate profile.}
    \label{fig:real1}
\end{figure*}

\subsection{Simulation and Real-world Experiments}

\begin{figure}[t]
    \centering
    \includegraphics[width=0.95\linewidth]{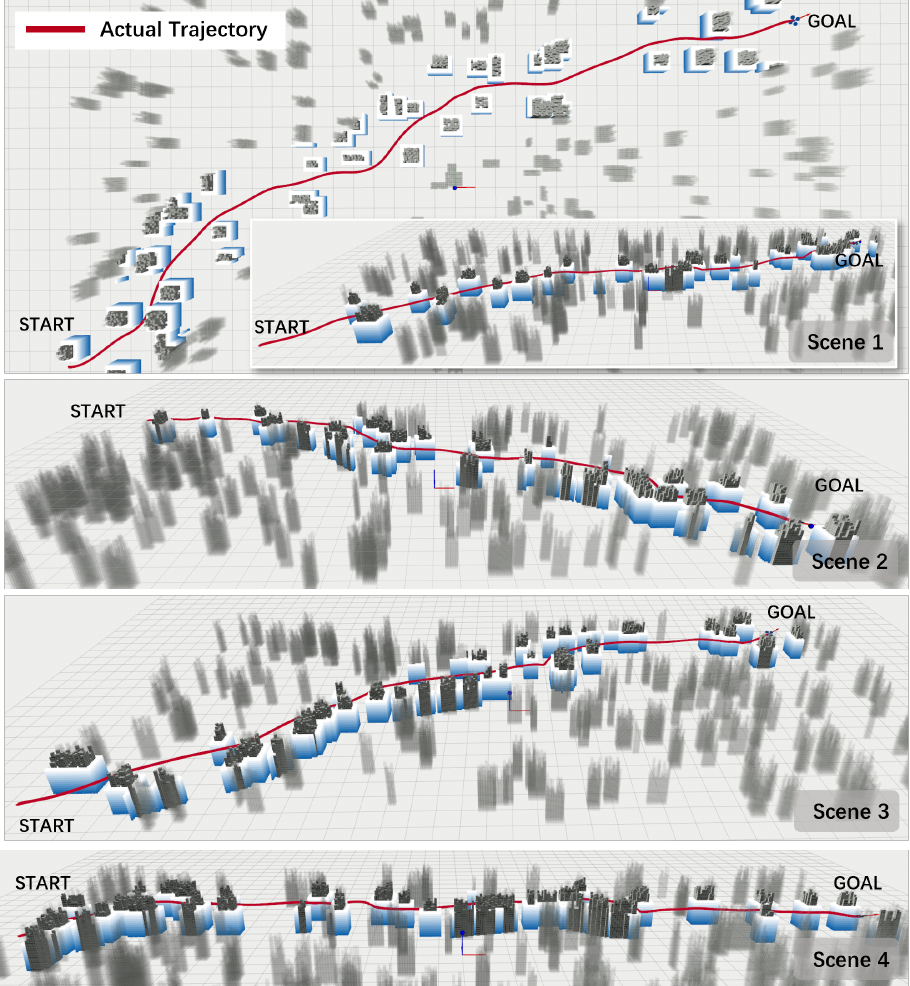}
    \caption{Visualization of the actual trajectory driven by the RL policy in the simulation environment.}
    \label{fig:rvizsim}
\end{figure}

\begin{figure}[t]
    \centering
    \includegraphics[width=0.9\linewidth]{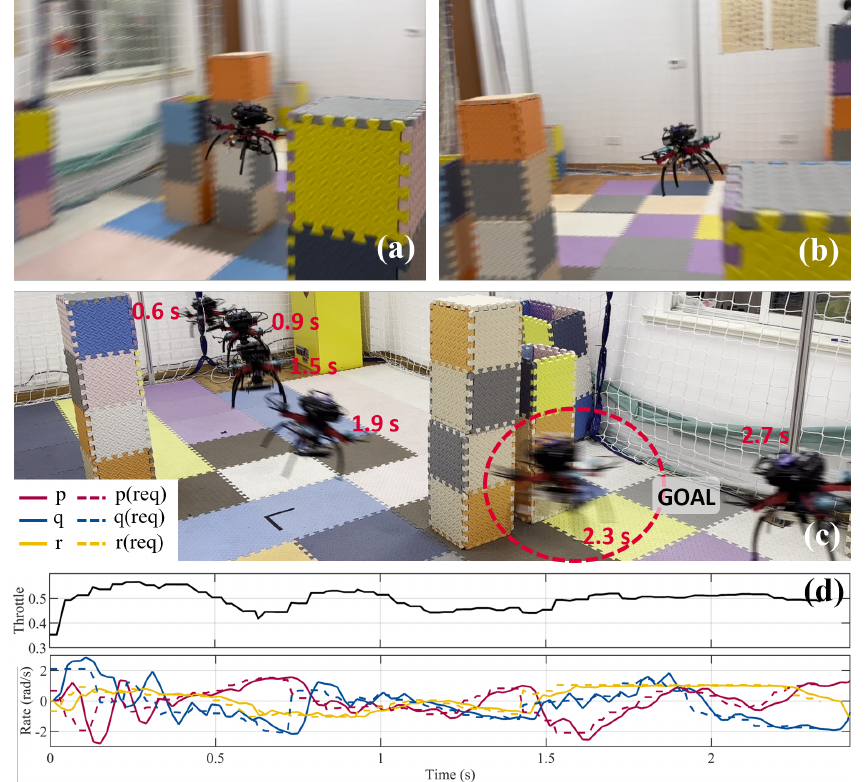}
    \caption{Real-world experiment conducted in an indoor environment: (a) The quadrotor successfully passes the obstacle. 
    (b) The quadrotor adjusts its attitude to navigate toward the goal. 
    (c) Composite image showing the sequence of the flight.
    (d) Desired throttle, rate and actual rate of the quadrotor.}
    \label{fig:real2}
\end{figure}

\begin{figure}[t]
    \centering
    \includegraphics[width=0.9\linewidth]{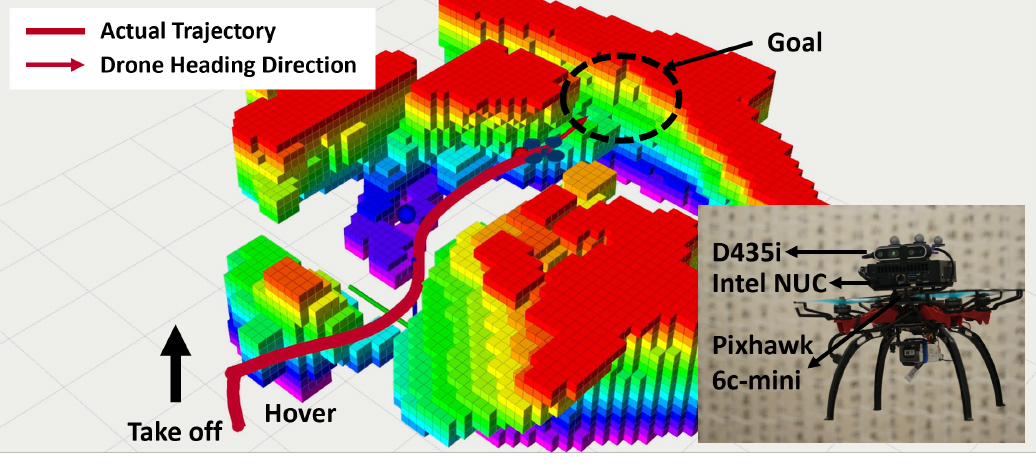}
    \caption{Trajectory of an indoor experiment and our quadrotor.}
    \label{fig:real3}
\end{figure}

We present several experiments conducted in both simulation and real-world environments to evaluation the feasibility and performance of the proposed method.
In simulation, we focus on navigating a quadrotor through some cluttered and randomized environments, where the RL policy generates control inputs that drive the quadrotor from the start point to the goal with control points provided by proposed searching method while avoiding obstacles.
In the simulation environment, as shown in Fig. \ref{fig:rvizsim}, the quadrotor follows control commands generated by RL policy, demonstrating the feasibility and real-time performance of our method.
The actual trajectory driven by the RL policy is visualized in red.

To validate the performance of proposed planning system in real-world conditions, we conducted experiments in an indoor environment.
The quadrotor's pose data is provided by the NOKOV motion capture system, and the ESDF map is generated using the pose and depth images.
As depicted in Fig. \ref{fig:real2}, the quadrotor successfully navigates around obstacles and adjusts its attitude to reach the goal. 
In Fig. \ref{fig:real3}, we show the actual trajectory of the quadrotor during a flight, further confirming the effectiveness of the RL policy in guiding the quadrotor to the goal while avoiding obstacles.

These experiments demonstrate the robustness and adaptability of the RL policy across both simulated and real-world environments, highlighting its potential for practical deployment in autonomous quadrotor motion planning tasks.

\section{Conclusion}

In this letter, we present an improved framework for quadrotor local planning that integrates path searching and control generation using visibility-based methods and reinforcement learning.
Our approach utilizes visibility path searching to generate safe control points with minimal path length, which are subsequently optimized by an RL policy to produce desired rate and throttle commands.
We validate our method in several simulated environments and compare its performance against baseline approaches, showing superior time efficiency.
We further conduct real-world experiments to demonstrate the feasibility and effectiveness of our method in practical applications.

Future work will focus on optimizing the RL policy to reduce energy consumption, enhance risk awareness through active yaw planning, and improve the success rate.
Additionally, we aim to extend our method to outdoor real-world scenarios and expand it to fully 3D space.
We also plan to investigate the potential of our approach in highly dynamic environments.

\bibliographystyle{IEEEtran}
\bibliography{main}

\begin{thebibliography}{10}
\providecommand{\url}[1]{#1}
\csname url@rmstyle\endcsname
\providecommand{\newblock}{\relax}
\providecommand{\bibinfo}[2]{#2}
\providecommand\BIBentrySTDinterwordspacing{\spaceskip=0pt\relax}
\providecommand\BIBentryALTinterwordstretchfactor{4}
\providecommand\BIBentryALTinterwordspacing{\spaceskip=\fontdimen2\font plus
\BIBentryALTinterwordstretchfactor\fontdimen3\font minus \fontdimen4\font\relax}
\providecommand\BIBforeignlanguage[2]{{%
\expandafter\ifx\csname l@#1\endcsname\relax
\typeout{** WARNING: IEEEtran.bst: No hyphenation pattern has been}%
\typeout{** loaded for the language `#1'. Using the pattern for}%
\typeout{** the default language instead.}%
\else
\language=\csname l@#1\endcsname
\fi
#2}}

\bibitem{10478625}
H.~Wang, H.~Li, B.~Zhou, F.~Gao, and S.~Shen, ``Impact-aware planning and control for aerial robots with suspended payloads,'' \emph{IEEE Transactions on Robotics}, vol.~40, pp. 2478--2497, 2024.

\bibitem{ietgaofei}
\BIBentryALTinterwordspacing
L.~Quan, L.~Han, B.~Zhou, S.~Shen, and F.~Gao, ``Survey of uav motion planning,'' \emph{IET Cyber-Systems and Robotics}, vol.~2, no.~1, pp. 14--21, 2020. [Online]. Available: \url{https://ietresearch.onlinelibrary.wiley.com/doi/abs/10.1049/iet-csr.2020.0004}
\BIBentrySTDinterwordspacing

\bibitem{timealloc}
F.~Gao, W.~Wu, J.~Pan, B.~Zhou, and S.~Shen, ``Optimal time allocation for quadrotor trajectory generation,'' in \emph{2018 IEEE/RSJ International Conference on Intelligent Robots and Systems (IROS)}, 2018, pp. 4715--4722.

\bibitem{rl-racing}
Y.~Song, M.~Steinweg, E.~Kaufmann, and D.~Scaramuzza, ``Autonomous drone racing with deep reinforcement learning,'' in \emph{2021 IEEE/RSJ International Conference on Intelligent Robots and Systems (IROS)}, 2021, pp. 1205--1212.

\bibitem{minimumsnap}
D.~Mellinger and V.~Kumar, ``Minimum snap trajectory generation and control for quadrotors,'' in \emph{2011 IEEE International Conference on Robotics and Automation}, 2011, pp. 2520--2525.

\bibitem{safecorridor}
S.~Liu, M.~Watterson, K.~Mohta, K.~Sun, S.~Bhattacharya, C.~J. Taylor, and V.~Kumar, ``Planning dynamically feasible trajectories for quadrotors using safe flight corridors in 3-d complex environments,'' \emph{IEEE Robotics and Automation Letters}, vol.~2, no.~3, pp. 1688--1695, 2017.

\bibitem{gradbased}
F.~Gao, Y.~Lin, and S.~Shen, ``Gradient-based online safe trajectory generation for quadrotor flight in complex environments,'' in \emph{2017 IEEE/RSJ International Conference on Intelligent Robots and Systems (IROS)}, 2017, pp. 3681--3688.

\bibitem{bsplineframework}
W.~Ding, W.~Gao, K.~Wang, and S.~Shen, ``An efficient b-spline-based kinodynamic replanning framework for quadrotors,'' \emph{IEEE Transactions on Robotics}, vol.~35, no.~6, pp. 1287--1306, 2019.

\bibitem{fastmarching}
F.~Gao, W.~Wu, Y.~Lin, and S.~Shen, ``Online safe trajectory generation for quadrotors using fast marching method and bernstein basis polynomial,'' in \emph{2018 IEEE International Conference on Robotics and Automation (ICRA)}, 2018, pp. 344--351.

\bibitem{fastplanner}
B.~Zhou, F.~Gao, L.~Wang, C.~Liu, and S.~Shen, ``Robust and efficient quadrotor trajectory generation for fast autonomous flight,'' \emph{IEEE Robotics and Automation Letters}, vol.~4, no.~4, pp. 3529--3536, 2019.

\bibitem{faster}
J.~Tordesillas, B.~T. Lopez, M.~Everett, and J.~P. How, ``Faster: Fast and safe trajectory planner for navigation in unknown environments,'' \emph{IEEE Transactions on Robotics}, vol.~38, no.~2, pp. 922--938, 2022.

\bibitem{eth2016}
H.~Oleynikova, M.~Burri, Z.~Taylor, J.~Nieto, R.~Siegwart, and E.~Galceran, ``Continuous-time trajectory optimization for online uav replanning,'' in \emph{2016 IEEE/RSJ International Conference on Intelligent Robots and Systems (IROS)}, 2016, pp. 5332--5339.

\bibitem{fastplanner2}
B.~Zhou, F.~Gao, J.~Pan, and S.~Shen, ``Robust real-time uav replanning using guided gradient-based optimization and topological paths,'' in \emph{2020 IEEE International Conference on Robotics and Automation (ICRA)}, 2020, pp. 1208--1214.

\bibitem{raptor}
B.~Zhou, J.~Pan, F.~Gao, and S.~Shen, ``Raptor: Robust and perception-aware trajectory replanning for quadrotor fast flight,'' \emph{IEEE Transactions on Robotics}, vol.~37, no.~6, pp. 1992--2009, 2021.

\bibitem{egoplanner}
X.~Zhou, Z.~Wang, H.~Ye, C.~Xu, and F.~Gao, ``Ego-planner: An esdf-free gradient-based local planner for quadrotors,'' \emph{IEEE Robotics and Automation Letters}, vol.~6, no.~2, pp. 478--485, 2021.

\bibitem{minco}
Z.~Wang, X.~Zhou, C.~Xu, and F.~Gao, ``Geometrically constrained trajectory optimization for multicopters,'' \emph{IEEE Transactions on Robotics}, vol.~38, no.~5, pp. 3259--3278, 2022.

\bibitem{sci-wildflight}
\BIBentryALTinterwordspacing
A.~Loquercio, E.~Kaufmann, R.~Ranftl, M.~Müller, V.~Koltun, and D.~Scaramuzza, ``Learning high-speed flight in the wild,'' \emph{Science Robotics}, vol.~6, no.~59, p. eabg5810, 2021. [Online]. Available: \url{https://www.science.org/doi/abs/10.1126/scirobotics.abg5810}
\BIBentrySTDinterwordspacing

\bibitem{rl-mintimeflight}
R.~Penicka, Y.~Song, E.~Kaufmann, and D.~Scaramuzza, ``Learning minimum-time flight in cluttered environments,'' \emph{IEEE Robotics and Automation Letters}, vol.~7, no.~3, pp. 7209--7216, 2022.

\bibitem{rl-pawareflight}
Y.~Song, K.~Shi, R.~Penicka, and D.~Scaramuzza, ``Learning perception-aware agile flight in cluttered environments,'' in \emph{2023 IEEE International Conference on Robotics and Automation (ICRA)}, 2023, pp. 1989--1995.

\bibitem{sci-rlracing}
\BIBentryALTinterwordspacing
Y.~Song, A.~Romero, M.~Müller, V.~Koltun, and D.~Scaramuzza, ``Reaching the limit in autonomous racing: Optimal control versus reinforcement learning,'' \emph{Science Robotics}, vol.~8, no.~82, p. eadg1462, 2023. [Online]. Available: \url{https://www.science.org/doi/abs/10.1126/scirobotics.adg1462}
\BIBentrySTDinterwordspacing

\bibitem{10582409}
G.~Zhao, T.~Wu, Y.~Chen, and F.~Gao, ``Learning speed adaptation for flight in clutter,'' \emph{IEEE Robotics and Automation Letters}, vol.~9, no.~8, pp. 7222--7229, 2024.

\bibitem{JPS}
\BIBentryALTinterwordspacing
D.~Harabor and A.~Grastien, ``Online graph pruning for pathfinding on grid maps,'' \emph{Proceedings of the AAAI Conference on Artificial Intelligence}, vol.~25, no.~1, pp. 1114--1119, Aug. 2011. [Online]. Available: \url{https://ojs.aaai.org/index.php/AAAI/article/view/7994}
\BIBentrySTDinterwordspacing

\bibitem{bresenham}
J.~E. Bresenham, ``Algorithm for computer control of a digital plotter,'' \emph{IBM Systems Journal}, vol.~4, no.~1, pp. 25--30, 1965.

\bibitem{hybridAstar}
\BIBentryALTinterwordspacing
D.~Dolgov, S.~Thrun, M.~Montemerlo, and J.~Diebel, ``Path planning for autonomous vehicles in unknown semi-structured environments,'' \emph{The International Journal of Robotics Research}, vol.~29, no.~5, pp. 485--501, 2010. [Online]. Available: \url{https://doi.org/10.1177/0278364909359210}
\BIBentrySTDinterwordspacing

\bibitem{benchmark-action}
E.~Kaufmann, L.~Bauersfeld, and D.~Scaramuzza, ``A benchmark comparison of learned control policies for agile quadrotor flight,'' in \emph{2022 International Conference on Robotics and Automation (ICRA)}, 2022, pp. 10\,504--10\,510.

\bibitem{openaigym}
\BIBentryALTinterwordspacing
G.~Brockman, V.~Cheung, L.~Pettersson, J.~Schneider, J.~Schulman, J.~Tang, and W.~Zaremba, ``Openai gym,'' 2016. [Online]. Available: \url{https://arxiv.org/abs/1606.01540}
\BIBentrySTDinterwordspacing

\bibitem{PPO}
\BIBentryALTinterwordspacing
J.~Schulman, F.~Wolski, P.~Dhariwal, A.~Radford, and O.~Klimov, ``Proximal policy optimization algorithms,'' 2017. [Online]. Available: \url{https://arxiv.org/abs/1707.06347}
\BIBentrySTDinterwordspacing

\bibitem{stable-baselines3}
\BIBentryALTinterwordspacing
A.~Raffin, A.~Hill, A.~Gleave, A.~Kanervisto, M.~Ernestus, and N.~Dormann, ``Stable-baselines3: Reliable reinforcement learning implementations,'' \emph{Journal of Machine Learning Research}, vol.~22, no. 268, pp. 1--8, 2021. [Online]. Available: \url{http://jmlr.org/papers/v22/20-1364.html}
\BIBentrySTDinterwordspacing

\end{thebibliography}

\addtolength{\textheight}{-12cm}   

\end{document}